\documentclass[11pt,a4paper]{article}
\usepackage[table,xcdraw]{xcolor}
\usepackage[hyperref]{acl2021}
\usepackage{times}
\usepackage{latexsym}
\usepackage{graphicx}
\usepackage{booktabs}

\usepackage{microtype}

 \aclfinalcopy 

\title{Are Mutually Intelligible Languages Easier to Translate?}

\author{
Avital Friedland\thanks{~~Equal contribution.} \qquad Jonathan Zeltser$^*$ \qquad Omer Levy \\
Tel Aviv University \\
\texttt{\{avitalfried,jonathanz1\}@mail.tau.ac.il}
}

\date{}

\begin{document}
\maketitle

\begin{abstract}
Two languages are considered mutually intelligible if their native speakers can communicate with each other, while using their own mother tongue.
How does the fact that humans perceive a language pair as mutually intelligible affect the ability to learn a translation model between them?
We hypothesize that the amount of data needed to train a neural machine translation model is anti-proportional to the languages' mutual intelligibility.
Experiments on the Romance language group reveal that there is indeed strong correlation between the area under a model's learning curve and mutual intelligibility scores obtained by studying human speakers.
\end{abstract}
\section{Introduction}

When speakers of two different languages can effectively communicate each in their own tongue, these languages are considered \textit{mutually intelligible}.
Intelligibility is often assymetric, and considered to be a continuous notion \cite{gooskens}, with some languages exhibiting higher intelligibility (e.g. Bulgarian and Macedonian) while others are only partially understood in the oral form (German and Yiddish) or written media (Russian and Ukrainian).
Does mutual intelligibility, as perceived by humans, make translation an easier task for machines to learn?

We hypothesize that mutually intelligible language pairs require less bitext data to train a neural machine translation model from one language to the other.
To test our hypothesis, we train machine translation models on varying amounts of data for language pairs in the Romance language group (Spanish, Portuguese, Italian, French, and Romanian).
For each language pair, we produce a learning curve that measures the relative BLEU score (with respect to the model trained on 100\% of the data), and compute the area under the curve (AUC), which we interpret as how ``easy'' it is for a machine to learn to translate this particular language pair.

We compare the AUC with two intelligibility scores: an automatically-computed metric of language similarity based on lexical analysis \cite{dinu-ciobanu-2014-romance}, and a human measure of intelligibility derived from a spoken cloze test \cite{gooskens}.
The lexical metric shows high correlation with the AUC (Pearson's $r = 0.539$), while the spoken-language metric exhibits lower correlation ($r = 0.224$).
Further investigation reveals that due to cultural exposure, Romanian speakers find Italo-Western languages intelligible, but not vice versa.
Treating these data points as outliers reveals strong correlation between the spoken-language metric and the AUC ($r = 0.585$),
indicating that translating between languages that humans perceive to be mutually intelligible may indeed be easier.

\section{Measuring Mutual Intelligibility}
\label{sec:background}

There are various methods in applied linguistics for testing to what degree a speaker of one language (source) can understand utterances from a related language (target).
In this work, we focus on five Romance languages (Spanish, Portuguese, Italian, French, and Romanian) and rely on two different approaches for measuring their mutual intelligibility.
\citet{dinu-ciobanu-2014-romance} use automatic lexical analysis to find etymons and cognates (i.e. words that are genetically related) to measure language similarity.
Table~\ref{tab:dinu} shows their intelligibility scores computed over the Europarl corpus \cite{europarl}.
We use this metric as a proxy for written mutual intelligibility, with the significant caveat that it is not based on a study with human participants.

For a more accurate measure of human-perceived mutual intelligibility, we turn to a recent study by \citet{gooskens}.
Here, spoken language intelligibility is measured using a cloze test, in which selected words are removed from a target language utterance and replaced by beeps of uniform length.
The source language speaker must then fill in the missing words; source to target intelligibility is thus how \textit{accurately} the average subject does so.
Since previous exposure to the target language via education, culture, and media can significantly skew the metric, Gooskens et al. select participants with minimal exposure to the target language a priori.
Table~\ref{tab:gooskens} shows their spoken language intelligibility scores among the five Romance languages.

\begin{table}[t]
\small
\centering
\begin{tabular}{lccccc}
\toprule
\textbf{src{\textbackslash}tgt} & \textbf{es} & \textbf{pt} & \textbf{it} & \textbf{fr} & \textbf{ro} \\
\midrule
\textbf{es} & --- & 86.40 & 77.37 & 68.86 & 59.97 \\
\textbf{pt} & 84.56 & --- & 75.33 & 67.11 & 64.54 \\
\textbf{it} & 79.45 & 75.65 & --- & 68.82 & 65.97 \\
\textbf{fr} & 69.98 & 63.15 & 72.93 & --- & 65.12 \\
\textbf{ro} & 70.17 & 65.57 & 69.15 & 62.27 & --- \\ 
\bottomrule
\end{tabular}
\caption{Written language intelligibility, based on automatic lexical analysis \cite{dinu-ciobanu-2014-romance}.}
\label{tab:dinu}
\end{table}

\begin{table}[t]
\small
\centering
\begin{tabular}{lccccc}
\toprule
\textbf{src{\textbackslash}tgt} & \textbf{es} & \textbf{pt} & \textbf{it} & \textbf{fr} & \textbf{ro} \\
\midrule
\textbf{es} & --- & 35.7 & 38.2 & 28.2 & 13.7 \\
\textbf{pt} & 62.0 & --- & 44.1 & 34.3 & 14.7 \\
\textbf{it} & 56.0 & 23.4 & --- & 18.6 & ~~8.7 \\
\textbf{fr} & 31.5 & 23.5 & 22.9 & --- & 11.0 \\
\textbf{ro} & 46.6 & 20.7 & 47.2 & 47.1 & --- \\
\bottomrule
\end{tabular}
\caption{Spoken language intelligibility, based on human subject spoken cloze tests \cite{gooskens}.}
\label{tab:gooskens}
\end{table}

\section{Experiment}

We hypothesize that mutually intelligible language pairs require less data to train machine translation models than language pairs that exhibit less intelligibility among human speakers.
To test our hypothesis, we sample the training set of each language pair to create training subsets of various sizes, and use these training subsets to create a learning curve for each language pair.
We then compare the area under the learning curves (AUC) to mutual intelligibility scores (Table~\ref{tab:dinu} and Table~\ref{tab:gooskens}); if the hypothesis is correct, we expect to observe correlation between AUC and intelligibility scores.

\subsection{Setup}

\paragraph{Data}
To leverage existing research on mutual intelligibility (Section~\ref{sec:background}), we focus on translation among five Romance languages: Spanish (es), Portuguese (pt), Italian (it), French (fr), and Romanian (ro).
We use the IWSLT'14 shared task \cite{cettolo2014report}, which consists of TED talk subtitles translated from the original spoken English to many other languages, to construct our training data.
Specifically, we pivot through English, enumerating over each sentence in the source language, and matching its translation in the target language by searching for the identical pivot sentence in English.
This process creates equivalent parallel data for all 20 language pairs.

\paragraph{Model and Hyperparameters}
We use fairseq \cite{ott2019fairseq} to train a small transformer model \cite{transformers} with 6 layers of encoder/decoder, 256 model dimensions, 1024 hidden dimensions, and 4 attention heads per layer.
Each model was trained for 50 epochs with dynamic batches of 8000 max tokens on a single 12GB GPU.
We use the Adam optimizer with inverse square root learning rate scheduling, warming up for 4000 steps and peaking at a learning rate of 0.0005, and regularize with 0.3 dropout, smoothed cross entropy loss, weight decay of 0.0001.
The best checkpoint is selected via validation set loss.
During evaluation, we decode with beam search (beam 5) and evaluate with SacreBLEU \cite{post-2018-call}.

\subsection{Measuring Translation Difficulty}

We describe visual and quantitative means for measuring ``learnability'' by plotting the relative performance (BLEU score) as a function of the number of examples provided at training time.

\paragraph{Visual Measure: Learning Curve}
We subsample the original training data to create smaller training subsets, ranging from 20\% of to the original data to 100\%, in increments of 10\%.
For each subset, we train a model and record its BLEU score on the test data (which was not modified).
Since comparing BLEU scores between different language pairs is not informative, we compute the \textit{relative} BLEU score by dividing each value by the score achieved by the model trained on 100\% of the training data.\footnote{For example, if a model trained on half the data achieved 24 BLEU while a model trained on all the data achieved 30 BLEU, their relative performance measure would be 0.8 for the half-data model and 1.0 for the full-data model.}
Using relative performance to plot learning curves allows us to compare learning curves of different language pairs.

\begin{figure}
\centering
\includegraphics[width=\columnwidth,height=\textheight,keepaspectratio]{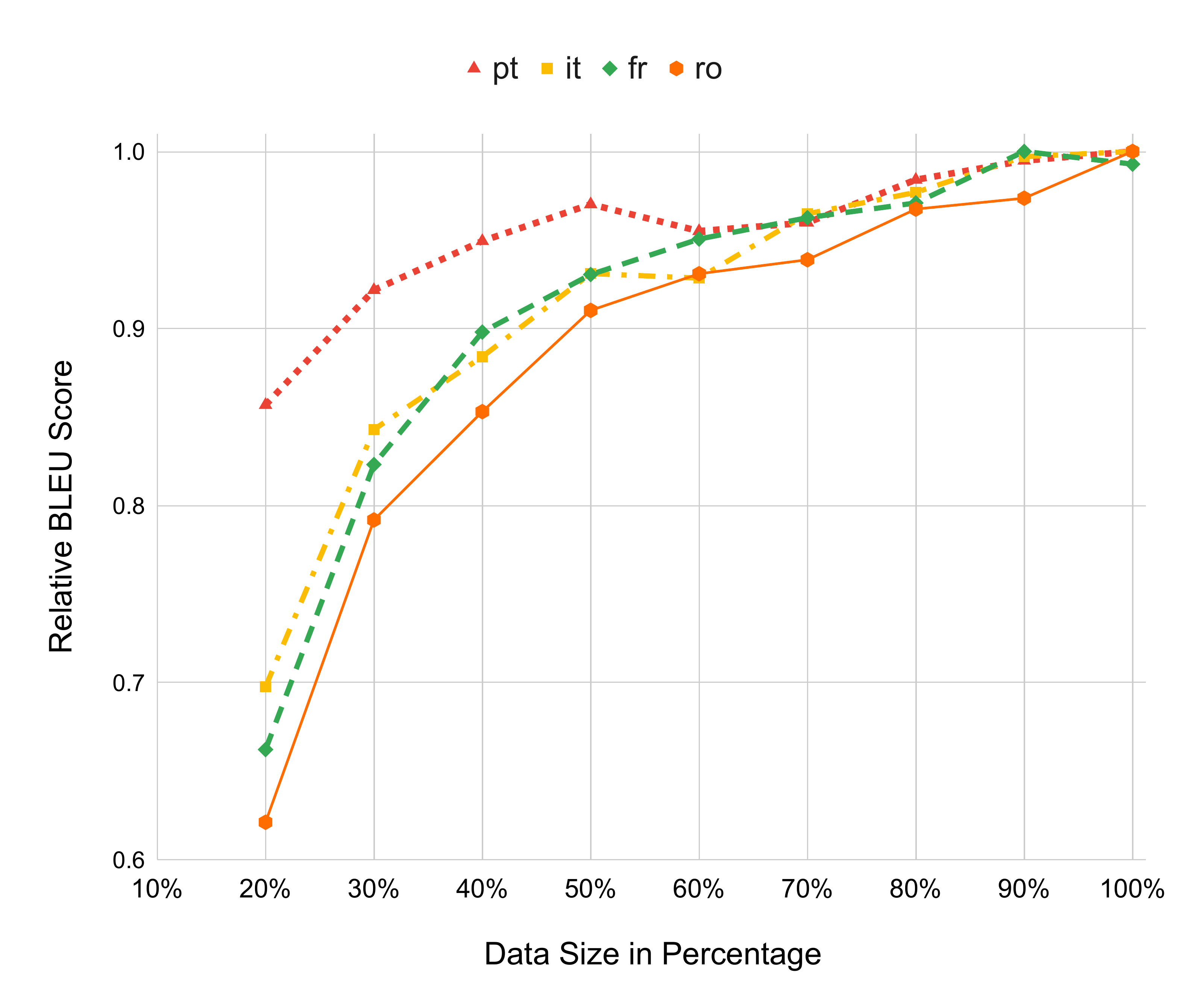}
\caption{Machine translation learning curves from Spanish (es) to other Romance languages. The horizontal axis reflects the percentage of training data used. The vertical axis is the BLEU score on the test set, normalized against the BLEU score of the full-data model.}
\label{fig:es_to_all}
\end{figure}

Figure~\ref{fig:es_to_all} presents learning curves for translation from Spanish (es) to other Romance languages.
We observe that generally, the Spanish - Portuguese pair learns quickest (i.e., reaches high relative BLEU score with smallest percentage of data used). We can also observe that the initial relative BLEU scores are in the same order as in the scores in both the lexical analysis (Table \ref{tab:dinu}) and spoken analysis (Table \ref{tab:gooskens}). This congruence in ordering is not preserved entirely as percentage of data increases, but is the most prevalent ordering.

\paragraph{Quantitative Measure: Area Under Curve}
The learning curves visualize how fast the performance saturates when the model is provided with more training data.
For a numerical measure of this phenomenon, we compute the area under the curve (AUC) for each language pair, using the trapezoidal rule.
We interpret AUC as our measure for how ``easy'' it is for a machine to learn to translate a particular language pair.

Table~\ref{tab:auc} shows the AUC score for every language pair.
We observe that the lowest 40\% of the scores involve Romanian in the role of either source or target language.
Languages with high mutual intelligibility on the cloze test, such as Spanish and Portuguese also achieve the highest AUC.
From a manual comparison, these scores seem to align well with the intelligibility scores (Table~\ref{tab:dinu} and Table~\ref{tab:gooskens}), but a more rigorous method is needed to determine how strong this correlation really is.


\begin{table}[t!]
\small
\centering
\begin{tabular}{lccccc}
\toprule
\textbf{src{\textbackslash}tgt} & \textbf{es} & \textbf{pt} & \textbf{it} & \textbf{fr} & \textbf{ro} \\
\midrule
\textbf{es} & --- & 74.71 & 73.74 & 73.63 & 71.77 \\
\textbf{pt} & 75.12 & --- & 72.32 & 72.83 & 70.05 \\
\textbf{it} & 74.34 & 73.32 & --- & 72.69 & 71.71 \\
\textbf{fr} & 72.89 & 72.78 & 72.54 & --- & 69.94 \\
\textbf{ro} & 71.87 & 70.67 & 71.48 & 70.48 & --- \\ 
\bottomrule
\end{tabular}
\caption{Area under curve (AUC) of machine translation learning curves. More area implies saturation with less data.}
\label{tab:auc}
\end{table}

\begin{figure*}[t]
\centering
\includegraphics[width=\textwidth]{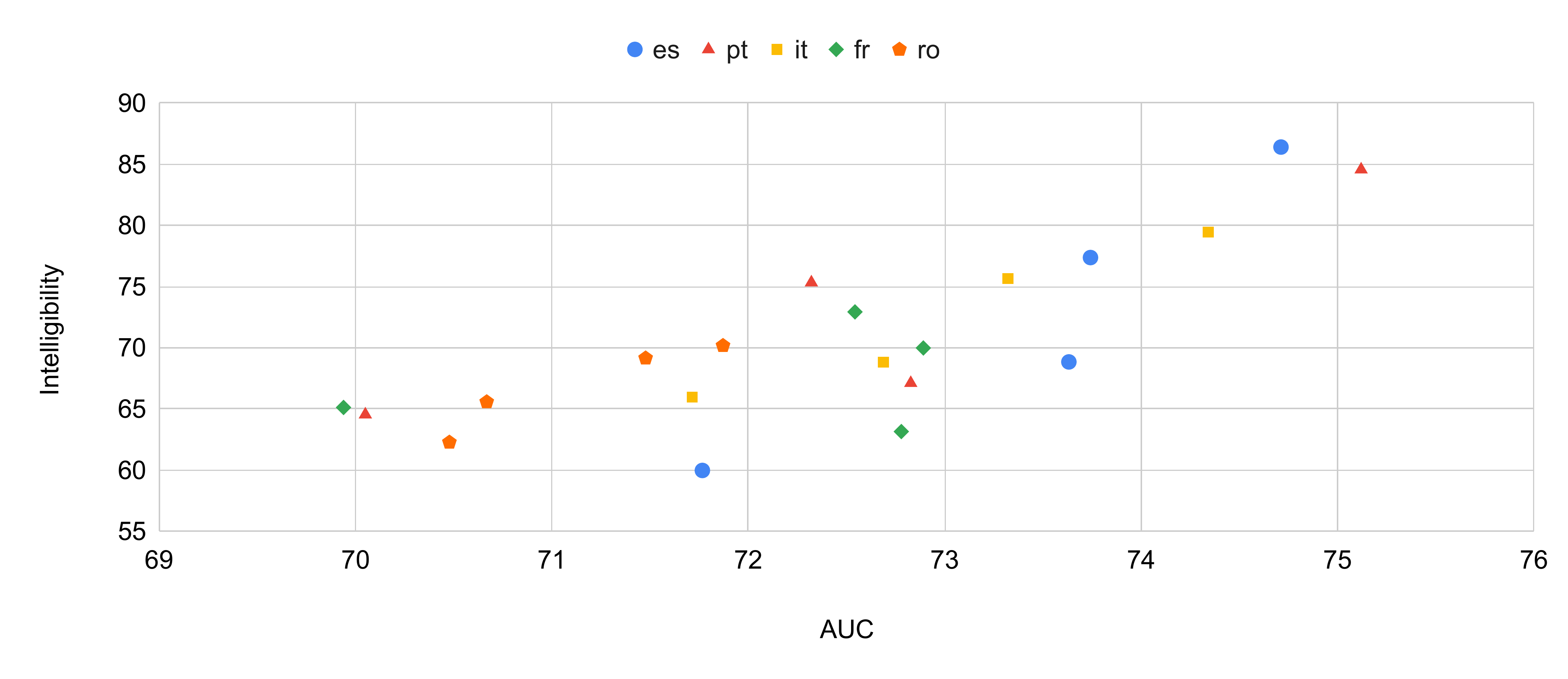}
\caption{Correlation between the area under the learning curve (AUC) and written-language mutual intelligibility based on automatic lexical analysis \cite{dinu-ciobanu-2014-romance}. Pearson's $r =  0.539$. Each color and shape of a data point indicate the data point's source language, both in the intelligibility analysis and the translation model.}
\label{fig:auc_dinu}
\end{figure*}

\begin{figure*}[t!]
\centering
\includegraphics[width=\textwidth]{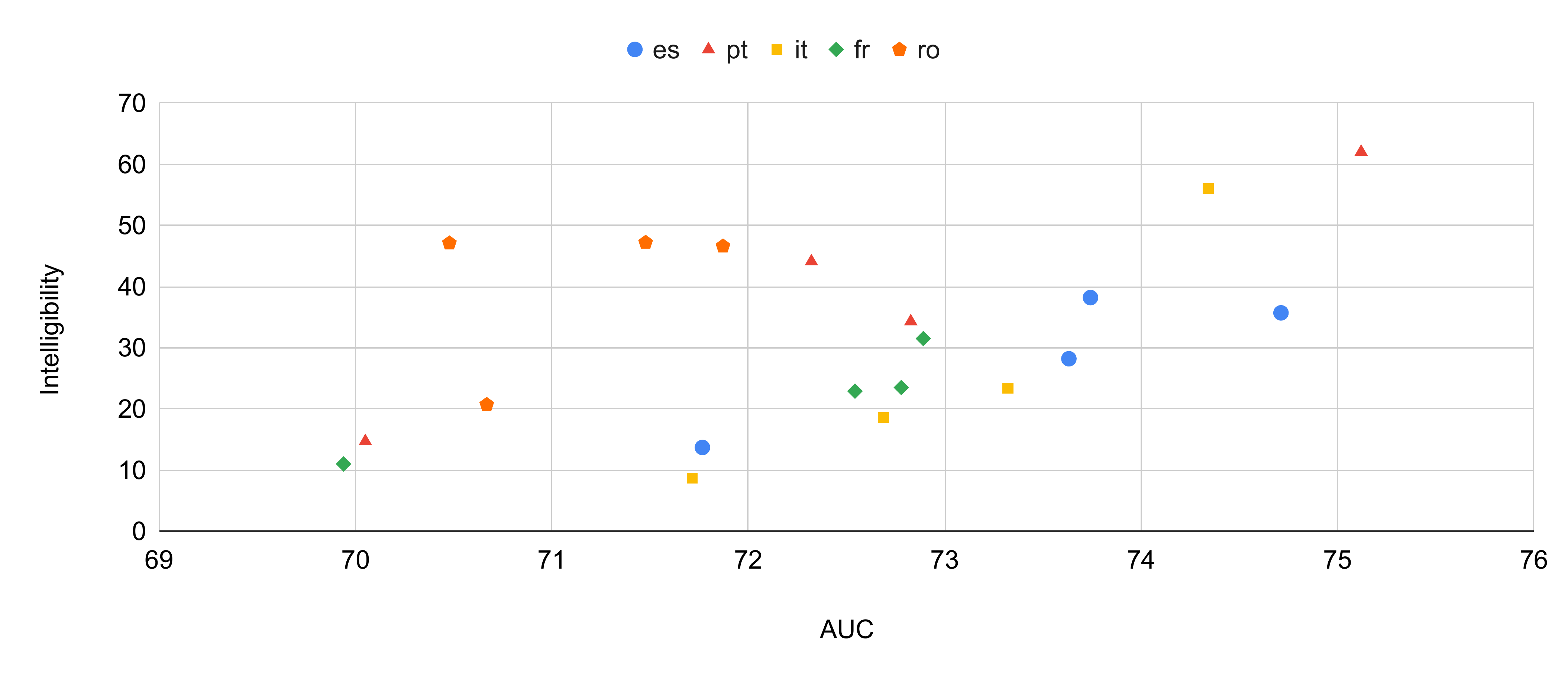}
\caption{Correlation between the area under the learning curve (AUC) and spoken-language mutual intelligibility based on cloze tests with human participants \cite{gooskens}. Datapoints with Romanian source (orange pentagons) appear to be outliers to the general trend. Pearson's correlation with full data: $r = 0.224$. Without Romanian as a source language: $r = 0.585$. Each color and shape of a data point indicate the data point's source language, both in the intelligibility analysis and the translation model.}
\label{fig:auc_gooskens}
\end{figure*}

\subsection{Correlation with Mutual Intelligibility}

We plot all 20 language pairs using AUC as the horizontal axis and the intelligibility scores presented in mutual intelligibility studies (Section~\ref{sec:background}) as the vertical axis.
This allows us to visually compare the two metrics, and determine whether a trend exists.
In addition, we compute Pearson's correlation between each pair of metrics to provide a quantitative measure of the correlation.

\paragraph{Written Language Intelligibility}
Figure~\ref{fig:auc_dinu} presents the correlation between AUC scores (Table~\ref{tab:auc}) and written-language intelligibility scores based on lexical analysis (Table~\ref{tab:dinu}).
We observe a strong linear trend with no substantial outliers, which also matches the computed correlation of $r = 0.539$.
While this finding strengthens our hypothesis, it does not completely tease apart mutual intelligibility from language relatedness; we thus compare the AUC scores to an additional intelligibility metric based on human participant data.

\paragraph{Spoken Language Intelligibility}
Figure~\ref{fig:auc_gooskens} shows the correlation between the AUC scores and the spoken-language mutual intelligibility based on cloze tests.
The correlation is modest ($r = 0.224$), possibly due to the different medium (speech versus text).
However, a closer look at the figure brings to attention three datapoints in the top left quadrant, where Romanian is the source language (and the target language is French, Italian, Spanish, from left to right).
These points seem to be outliers to an otherwise strong linear trend.
One possible explanation is that Romanians learn French at school, and it is therefore difficult to find Romanian participants with minimal exposure to other languages in the Romance language group \cite{gooskens}.
When removing Romanian as a source language from the data, we see a much stronger correlation coefficient of $r = 0.585$, similar to the correlation observed with the written-language metric.

\section{Conclusions and Future Work}

This work presents the hypothesis that mutually intelligible language pairs are easier for neural machine translation models to learn, in the sense that they require less training data.
We test the hypothesis by computing the correlation between a proxy measure for learnability (AUC) and pre-existing metrics of mutual intelligibility. 
Experiments on five languages in the Romance language group show strong correlation, thus supporting the hypothesis.

To the best of our knowledge, this is the first use of our proposed methodology.
Future work may be able to utilize it to conduct a more extensive investigation of how mutual intelligibility affects machine translation by exploring more language pairs and groups, examine correlations with additional intelligibility metrics, and use true bitext that does not pivot through English.
Our findings might also suggest that there are potential opportunities in the intersection of few-shot learning and cross-lingual transfer between mutually intelligible languages.

\bibliographystyle{acl_natbib}
\bibliography{anthology,references}


\end{document}